\pdfoutput=1
\documentclass[11pt]{article}

\usepackage[letterpaper,margin=1in]{geometry}
\usepackage[utf8]{inputenc}
\usepackage{graphicx}
\usepackage{amsmath}
\usepackage[numbers,sort&compress]{natbib}
\usepackage[colorlinks=true,linkcolor=blue,citecolor=blue,urlcolor=blue]{hyperref}

\hypersetup{
  pdftitle={Synthetic and Derived Training Images for Campus Waste Detection: A Multi-Seed Evaluation with YOLOv8n},
  pdfauthor={Ali Behbahani; Newsha Javanmardi; Shahriar Ahmed; Phouvadeth Vathana},
  pdfsubject={Multi-seed evaluation of synthetic and derived training images for YOLOv8n waste detection},
  pdfkeywords={waste detection, object detection, YOLOv8n, synthetic training data, domain shift, transfer learning, edge deployment}
}

\title{\bfseries Synthetic and Derived Training Images for Campus Waste Detection:\\
A Multi-Seed Evaluation with YOLOv8n}

\author{Ali Behbahani\thanks{Corresponding author:
\texttt{abehbaha@vols.utk.edu}}, Newsha Javanmardi, Shahriar Ahmed,\\
 and Phouvadeth Vathana\\[0.4em]
\small University of Tennessee, Knoxville, USA}
\date{}

\begin{document}
\maketitle

\begin{abstract}
\noindent
Incorrect disposal can contaminate campus recycling streams, and a bin-mounted
camera could provide feedback as an item is discarded. We evaluated whether
synthetic and derived images improve a YOLOv8n detector for this view. The real
dataset contained 148 campus photographs: 86 for training, 31 for validation,
and 31 for testing. Twelve joint-training configurations varied the amount and
source of added images. We repeated seven principal settings with four matched
seeds and computed bootstrap percentile intervals over those seeds. The
real-only model reached a mean mAP@0.5 of 0.691 [0.665, 0.722]. Background
replacement reduced the mean to 0.560 [0.499, 0.619], isolated-object images
gave 0.680 [0.644, 0.724], and the full augmentation pool gave 0.487 [0.438,
0.537]. We also tested hand-and-forearm composites because every real photo
showed a held object. Two cutouts in the initial composite set came from test
photographs, so we discarded that experiment, rebuilt the set with
training-split cutouts, and reran all four seeds. The corrected paired
difference was $+0.034$ [$-0.063$, $0.199$], which does not support a reliable
hand-composite effect. Single-seed transfer experiments produced
source-dependent rankings between joint mixing and sequential pretraining.
None of the evaluated configurations exceeded the real-only baseline. The
reported intervals quantify seed variation; the 31-photo test set remains too
small for strong class-specific conclusions.
\end{abstract}

\vspace{0.5em}

\section{Introduction}
\label{sec:intro}

Incorrectly sorted items can contaminate otherwise recoverable material and
increase the labor and energy required for downstream separation. Computer
vision has consequently become an active area of solid-waste research, with
systems developed for material recognition, automated sorting, and assistance
at disposal points~\cite{abdallah}. The application considered here is a
campus bin that provides guidance before an item enters the waste stream. A
small camera mounted near the opening could observe the presented object,
assign it to a disposal category, and return immediate feedback to the user.
This point-of-disposal setting differs from post-collection sorting because a
decision must be made from a brief, close-range view while the item is still in
a person's hand. It also places practical limits on model size, inference
time, and camera placement. A detector that works in this setting could move
classification upstream, where a correction is still possible, but only if
its training data represent the view that the installed camera will actually
encounter. The central problem is therefore not simply whether a detector can
recognize waste objects; it is whether a compact detector trained from limited
campus data can retain that ability under the visual conditions of disposal.

The deployment view differs systematically from the images used by many waste
classifiers. An object is close to the camera and normally held in a hand;
fingers can cover part of its boundary, packaging may be bent or crushed, and
the visible background includes floors, walls, bins, and other indoor clutter
under mixed lighting. Viewpoint and illumination also change across buildings
and collection sessions. Product photographs usually isolate an undamaged
item, outdoor litter datasets show objects after disposal, and recycling-plant
datasets emphasize material moving through a conveyor scene. None reproduces
the same combination of scale, hand occlusion, object deformation, and campus
context. This mismatch matters because a detector learns correlations from
the complete image rather than from the object crop alone. Building a large,
labeled collection for the exact camera position would be preferable, but it
requires repeated access to disposal locations, physical examples from each
class, and careful annotation of every bounding box. The verified real dataset
available for this study contains only 148 photographs, of which 86 are
available for training. That scale makes augmentation attractive while also
making the evaluation unusually sensitive to the composition of both the
training and test splits.

Synthetic and derived images offer a comparatively inexpensive way to expand
a scarce deployment dataset~\cite{shorten}. Detection pipelines have pasted
object crops into new scenes~\cite{cutpaste,georgakis,ghiasi}, generated entire
scenes through simulation or domain randomization
~\cite{richter,ros,tobin,domainrand}, and created transformed copies through
changes in lighting, noise, rotation, or geometry. These approaches can expose
a model to more pixel arrangements without collecting and labeling an equal
number of new photographs. Their value, however, depends on what variation is
actually introduced. A transformed image can increase the nominal sample
count while preserving the identity, shape, and capture artifacts of its
source object. A pasted object can add background diversity while creating
boundaries, scale relationships, or scene combinations that do not occur at
deployment. Simulation can provide extensive annotation but may introduce a
different appearance gap. Consequently, ``more images'' and ``more useful
training evidence'' are not equivalent. This distinction is central in our
added-data pool: its 695 images are derived from approximately 100 base
objects, so several nominally different images share the same physical object.
The experiments must therefore separate the amount of added data from its
source and must compare additions against real-only training rather than
assuming that a larger training set is inherently better.

Scene context creates a particularly relevant form of domain mismatch.
Detectors use surrounding information when localizing and classifying
objects~\cite{torralba,divvala}, and their predictions can change when an
otherwise familiar object is moved into an implausible scene~\cite{rosenfeld}.
Work on context-aware cut-and-paste augmentation likewise shows that placement
without regard to the receiving scene can erase the expected benefit
~\cite{dvornik}. The contrast in our data is visible and consistent: every real
campus photograph shows an object held during disposal, whereas the
background-replacement images present the object without a hand. This
difference motivates a controlled composite experiment in which the same
background-replacement image is evaluated with a hand-and-forearm cutout added
while the class label, background, object pixels, and bounding box remain
fixed. The intervention is useful because it targets a deployment-specific
visual cue, but its interpretation must remain narrow. Adding the cutout also
adds skin texture, new edges, and partial occlusion; it cannot isolate an
abstract semantic notion of ``being held.'' Cutout provenance is equally
important because any crop taken from a test photograph would leak test-set
pixels into training. We therefore report only the corrected experiment built
from training-split cutouts and retain the discarded result solely to document
the correction.

Small-data comparisons also require attention to randomness and statistical
scope. Detector training depends on initialization, minibatch order, and
stochastic augmentation, so a favorable result from one seed may disappear
when the same configuration is repeated~\cite{dietterich,bouthillier}. A fixed
test split creates a separate source of uncertainty: repeating training can
show whether a ranking is stable across seeds, but it cannot show how that
ranking would change if different photographs or physical objects were held
out. We address the first problem by repeating seven principal settings with
four matched seeds, presenting the individual runs, and computing percentile
intervals over seed variation. We address the second by limiting all claims to
the fixed 31-photo test set and by treating class-level patterns as tentative,
especially for glass, which appears in only two test photographs. Deployment
adds another constraint that accuracy alone cannot answer. The intended bin
computer is a low-cost single-board device, so an otherwise accurate model may
still be unsuitable if export, quantization, or CPU inference fails. Our edge
measurements therefore complement the detection study, although the
four-thread workstation proxy is explicitly not presented as a direct
Raspberry Pi benchmark.

\begin{figure}[!htb]
\centering
\includegraphics[width=\textwidth]{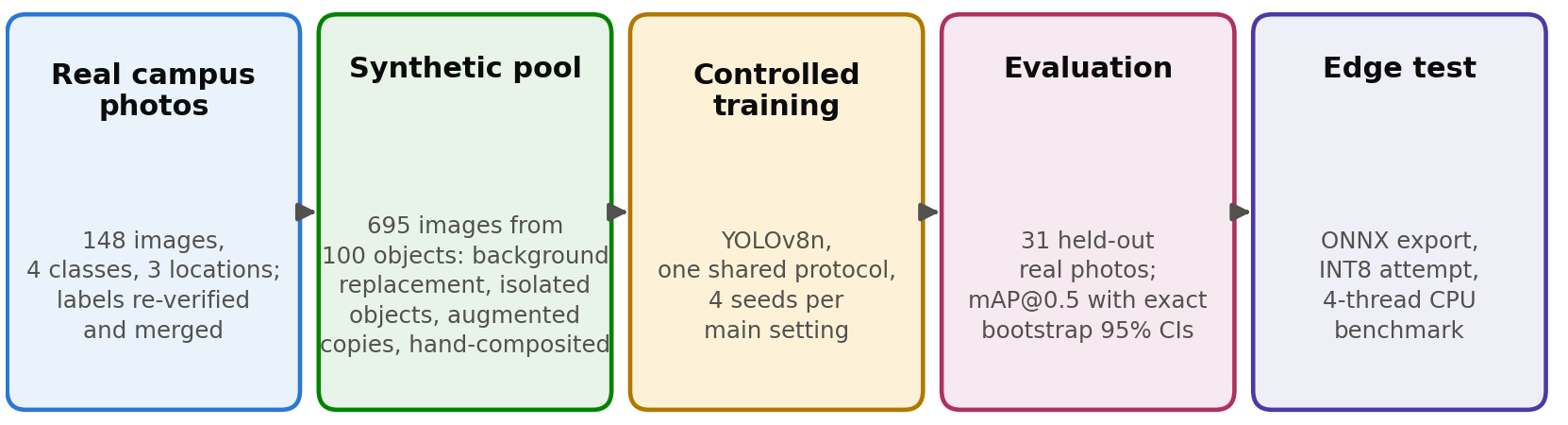}
\caption{Study design. We merged and verified three batches of real photos,
then trained YOLOv8n with different amounts and sources of added images. Every
model was evaluated on the same 31 real test photos. Selected models were also
exported for the CPU benchmark.}
\label{fig:system_overview}
\end{figure}

Against this background, the study holds the YOLOv8n architecture,
optimization protocol, validation procedure, and real test set fixed while
changing the training images. RQ1 asks whether either the amount or the source
of added data changes detection on real campus photographs. Its design includes
mixture ratios, reductions in the number of real training images, and
source-specific comparisons, with an approximately size-matched contrast
between isolated-object and background-replacement additions. RQ2 asks whether
the corrected hand-composite set changes performance relative to the same
background-replacement images without hands. RQ3 explores whether synthetic
images behave differently when mixed jointly with real photographs or used
for pretraining before real-image fine-tuning. Finally, the export study
examines FP32 models, an attempted INT8 conversion, and latency under a fixed
four-thread CPU setting. Together, these experiments are designed to identify
which conclusions the available data support and which remain limited by the
small test set, four-seed sample, or single-seed transfer runs. The study makes
five concrete contributions, summarized with the complete workflow in
Figure~\ref{fig:system_overview}:
\begin{enumerate}
\item verification and merging of a 148-photo, four-class campus dataset whose
three source batches used conflicting class indices;
\item twelve joint-training configurations that separate the amount and source
of added data, including four matched seeds for seven principal settings;
\item a corrected hand-composite experiment rebuilt with training-only
cutouts after test-split leakage was found in the initial cutout set;
\item a single-seed comparison of joint mixing and sequential pretraining for
three image sources; and
\item a four-thread CPU benchmark of the exported FP32 and INT8 models.
\end{enumerate}

\section{Related Work}
\label{sec:related}

\subsection{Waste Datasets and Smart Bins}
Public waste datasets differ from the smart-bin scene in systematic ways.
TrashNet shows single items on plain backgrounds~\cite{trashnet}, TACO
annotates outdoor litter lying in streets and parks~\cite{taco}, and WaRP and
ZeroWaste cover conveyor belts inside recycling
plants~\cite{warp,zerowaste}. These datasets do not target the specific view
of a hand presenting an item to a bin-mounted camera in a cluttered campus
interior. The difference is not a criticism of those resources; each dataset
was assembled for a different recognition problem and therefore encodes a
different relationship among object scale, background, deformation, and
camera position. It does mean that results obtained on isolated items,
post-disposal litter, or industrial material cannot be transferred directly
to point-of-disposal campus images. Smart-bin prototypes demonstrate the
practical appeal of local inference and have paired compact classifiers with
low-cost hardware. Jin et al., for example, built a sorting device around an
attention-augmented MobileNetV2 running on a Raspberry~Pi~\cite{jin2023}.
That line of work establishes the feasibility of a compact device but does not
resolve which training images best represent a held object at a campus bin.
Our study addresses the data side of that problem: it keeps the detector and
real test photographs fixed, then varies the quantity, visual source, and
integration of added images. This design makes the deployment mismatch itself
the object of evaluation rather than treating all waste imagery as
interchangeable.

\subsection{Object Detectors for Edge Deployment}
Modern detectors split into two-stage designs built around region
proposals~\cite{fasterrcnn} and one-stage designs that regress boxes
directly~\cite{yolo,ssd}. Resource-constrained systems often favor the latter.
One-stage models and compact backbones offer useful speed--accuracy
trade-offs~\cite{mobilenetv2}, which makes them plausible candidates for an
embedded bin controller. Architecture choice nevertheless introduces a major
confound in a data study: a larger backbone, a different detection head, or a
different pretraining source can change the result independently of the added
images. We therefore use one fixed detector, YOLOv8n~\cite{yolov8}, the nano
member of a widely used one-stage family, and initialize every run from the
same COCO-pretrained model. This decision does not establish that YOLOv8n is
the best possible detector for campus waste; it creates a controlled platform
for comparing training sets within a model small enough to motivate edge
testing. Deployment also depends on the exported representation. Post-training
integer quantization can reduce model size and, on compatible hardware,
accelerate inference~\cite{int8,krishnamoorthi}, but a successful conversion
does not guarantee numerically valid predictions or lower end-to-end latency.
We therefore report the attempted INT8 export, including its failure, alongside
FP32 size and timing in Section~\ref{sec:results-edge}. Holding architecture
and export settings fixed keeps the main comparison centered on training data
rather than model capacity.

\subsection{Synthetic Training Data}
Synthetic data for detection spans a spectrum of realism: game-engine and
simulator imagery~\cite{richter,ros}, domain randomization that deliberately
abandons realism~\cite{tobin,domainrand}, and cut-and-paste compositing of
real object crops onto real backgrounds~\cite{cutpaste,georgakis}. Compositing
has produced strong results for instance detection in several
settings~\cite{cutpaste,ghiasi}, but success depends on the objects, placement
rules, receiving scenes, and relationship between synthetic and target
domains. These distinctions are important for our pool, which combines base
objects, background replacement, and several transformations rather than a
single generative process. Treating every derived file as an independent new
example would overstate its diversity because multiple images preserve the
same underlying object. Dvornik et al.\ found that context-blind placement can
erase expected gains~\cite{dvornik}; that result motivates our matched
background-replacement and hand-composite comparison. Integration strategy is
another design choice. Joint mixing exposes the optimizer to real and added
images throughout training, whereas sequential transfer first fits the added
source and then adapts to real photographs. Hinterstoisser et al.\ protected
real-image features by freezing them during synthetic-data training
~\cite{hinter}. Our transfer experiment includes a related frozen-backbone
variant, but its order and data regime are different, so it is not a direct
replication. By testing source, amount, and integration separately where the
available pool permits, we evaluate specific augmentation choices rather than
synthetic data as a single undifferentiated category.

\subsection{Context in Object Detection}
Scene context changes both object localization and classification
~\cite{torralba,divvala}. Rosenfeld et al.\ made the failure visible by moving
familiar objects into unusual scenes~\cite{rosenfeld}. Our mismatch is more
specific. Every real object is held, whereas every background-replacement
object appears alone. A hand may provide a cue about scale and the action of
disposal, but it can also hide discriminative regions, introduce edges near a
bounding box, or become spuriously associated with a class. Context may help
when training and deployment scenes agree, yet the same dependence can become
a liability when those scenes differ. Context-aware augmentation methods try
to control that relationship by placing objects only where their surroundings
are plausible~\cite{dvornik}; our intervention instead holds each replacement
scene and object annotation fixed while adding a visible hand-and-forearm
cutout. That narrower comparison directly addresses the most consistent
visual difference in this dataset. It does not, however, isolate semantic
context in the causal sense. The composite adds pixels, texture, boundaries,
and occlusion at the same time, and a small library of hand cutouts can create
repeated patterns of its own. We consequently describe the experiment as a
hand-composite intervention, report the provenance correction in full, and
avoid claiming that any measured difference is caused by contextual meaning
alone.

\subsection{Reliable Comparison of Trained Models}
Single training runs are a noisy basis for comparing learning
pipelines~\cite{dietterich}, and much of the variance in modern benchmarks
comes from seeds and data sampling rather than the methods under
test~\cite{bouthillier}. This problem is amplified when the training and test
sets are small, because the apparent advantage of one configuration can depend
on a favorable initialization or sample order. We therefore repeat seven
principal settings with four matched seeds. Using the same seed set across
settings preserves a run-level pairing and allows differences to be calculated
within seed before they are summarized. The reported bootstrap intervals use
all resamples of the four observed runs~\cite{bootstrap}; they describe
variation among those runs, not uncertainty over new photographs, new data
splits, or the broader population of campus disposal events. Four observations
also provide very limited inferential resolution, and neither interval width
nor exclusion of zero should be read as a conventional population-level
significance test. We retain the individual seed values in the figures so that
readers can see whether a mean reflects consistent movement or one unusual
run. Section~\ref{sec:stats} defines the resampling procedure, paired
comparisons, and remaining sources of uncertainty explicitly.

\section{Dataset}
\label{sec:dataset}

\subsection{Real Photos and Label Verification}
\label{sec:dataset-audit}
\begin{figure}[!htb]
\centering
\includegraphics[width=\textwidth]{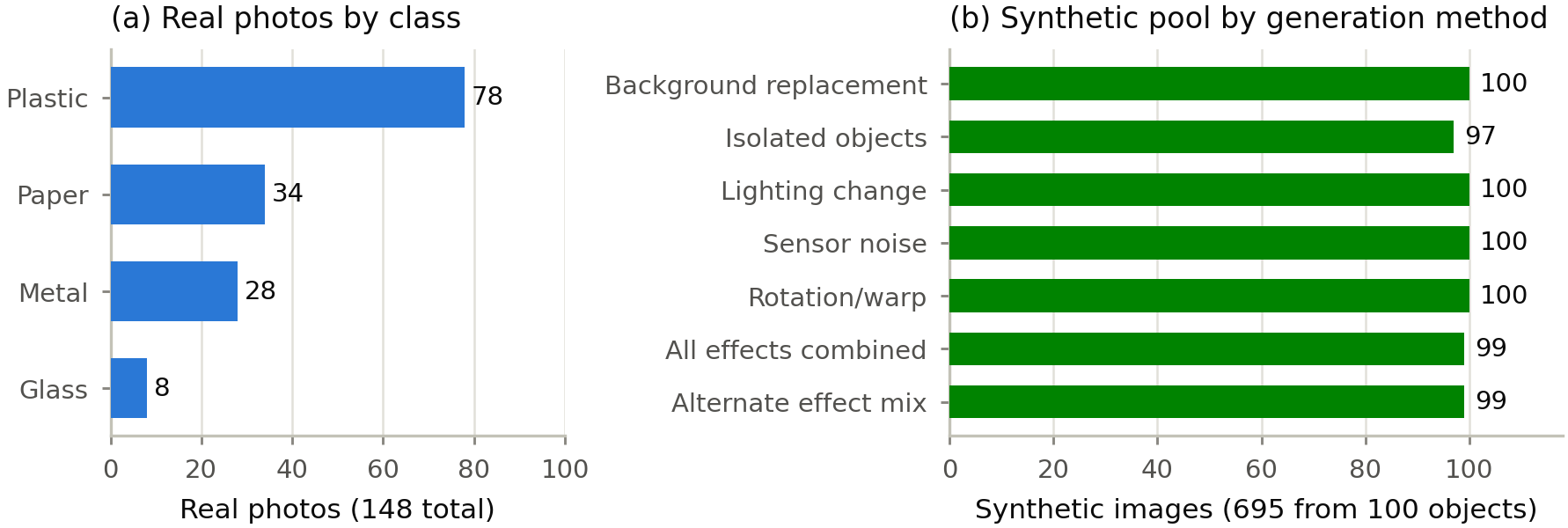}
\caption{Dataset composition. (a)~The 148 real photos are dominated by plastic;
glass is rare, which limits per-class conclusions for that class. (b)~The
synthetic/derived pool contains 695 images, but the generation methods
re-process the same 100 base objects (25 per class); image count therefore
overstates object diversity.}
\label{fig:dataset}
\end{figure}

The real data arrived in three batches of 35, 41, and 81 images collected at
three campus locations, and each batch assigned different numerical indices
to the four target classes. Before merging the batches, we compared each
configuration file with labeled sample images, established a common mapping,
and remapped every annotation to that index. We also removed a fifth,
out-of-scope ``food'' class. The verified dataset contains 148 photographs:
8~glass, 28~metal, 34~paper, and 78~plastic (Fig.~\ref{fig:dataset}a). We
generated one class-stratified split with random seed 2026. Images within each
class were shuffled and divided approximately 60/20/20, with rounding chosen
to preserve at least one validation and one test example per class. The final
allocation contains 86 training, 31 validation, and 31 test photographs, and
all four classes appear in each split. The test photographs were excluded from
training, validation, augmentation-source selection, and checkpoint selection.
Source-batch identifiers do not overlap, and the batches came from separate
sessions and locations, but the available metadata do not identify individual
physical objects within a batch. We also did not run an automated
near-duplicate search. The class mapping, label remapping, and class
stratification were therefore checked directly; group separation by physical
object remains unverified and is treated as a limitation rather than inferred
from the absence of matching filenames (Section~\ref{sec:limitations}).

\subsection{Synthetic and Derived Images}
\begin{figure}[!htb]
\centering
\includegraphics[width=\textwidth]{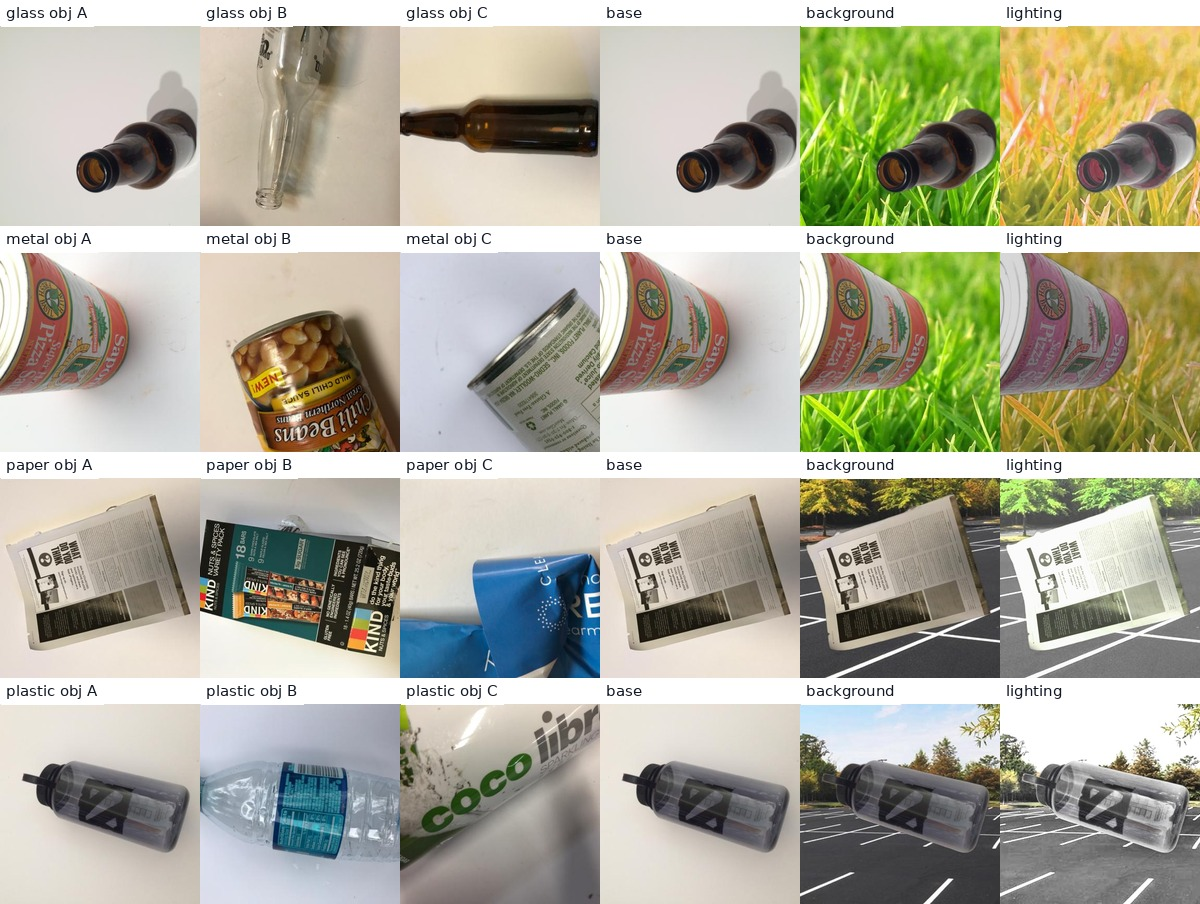}
\caption{Examples from the added-image pool. Columns 1--3 show three base
objects per class. Columns 4--6 follow one object through background
replacement and lighting transformations.}
\label{fig:synthetic_diversity}
\end{figure}

The added-data pool starts from 100 base images of distinct objects (25 per
class), each showing one item on a plain background. From each base image we
derived one background-replacement version (the object pasted into a new
scene) and up to five transformed copies using lighting changes, sensor noise,
rotation or warping, and two combined-effect pipelines. The 97 usable base
images form the isolated-object set, the type of object-instance collection
used in cut-and-paste augmentation~\cite{cutpaste}. We found no collection
record, license, or attribution that establishes who captured these images or
how they were created. Visual inspection cannot establish provenance, so we
refer to all 695 added images as the ``synthetic/derived pool'' and retain the
missing source record as a limitation. Figures~\ref{fig:dataset}b
and~\ref{fig:synthetic_diversity} show the method counts and representative
examples. The pool contains roughly seven images per base object; more images
do not add physical-object diversity. This distinction guides the subsequent
interpretation: sample count is reported exactly, but it is not used as a
surrogate for the number of independent object instances.

\section{Experimental Setup}
\label{sec:methods}

\subsection{Detector and Training Protocol}
\label{sec:protocol}
All runs use the nano variant of the You Only Look Once version~8 detector
(YOLOv8n)~\cite{yolov8}, chosen because the target bin computer needs a model
in the few-megabyte range. Each run starts from COCO-pretrained weights
~\cite{coco} and uses 640-pixel images, batch size 16, AdamW, automatic
learning-rate selection, and 100 epochs. The matched seeds are 2026, 1, 2,
and 3 for all seven principal settings (Section~\ref{sec:stats}). Training ran
on one NVIDIA GeForce RTX 3060 (12~GB) under Ubuntu 24.04.3 LTS.
The software environment comprised Python 3.12.9, PyTorch 2.7.0 (CUDA~12.6
build), and Ultralytics 8.4.76. We retained Ultralytics' \texttt{best.pt}
checkpoint for each run, selected by validation mAP@0.5:0.95, and evaluated it
once on the test split. The test results did not influence model selection.
Keeping that selection rule constant is important because the number and
source of training images are the experimental variables; changing the
architecture, initialization source, epoch limit, or checkpoint criterion
would introduce additional explanations for any difference. Every retained
model is evaluated on the same 31 held-out real photographs. We report mean
average precision (mAP) at an intersection-over-union (IoU) threshold of 0.5
(mAP@0.5) and averaged over IoU thresholds from 0.5 to 0.95
(mAP@0.5:0.95), computed with a confidence floor of 0.001 over the full
precision--recall curve~\cite{voc}. Precision and recall use a separate fixed
operating point: confidence 0.25 and non-maximum-suppression IoU 0.7. The
distinction matters because mAP summarizes a curve across confidence levels,
whereas a deployed bin controller must act on thresholded predictions. We use
the curve-based metric for the primary model comparison and the fixed
operating point only for the reported precision, recall, and qualitative
outcome counts.

\subsection{Multi-Seed Statistics}
\label{sec:stats}
Single training runs of a small detector on a small dataset are
noisy~\cite{dietterich,bouthillier}. We use four matched seeds. For each of the
seven principal settings, we form the reported 95\% percentile interval from
all $4^4=256$ bootstrap resamples of the four runs~\cite{bootstrap}. Paired
comparisons resample the four seed-wise differences. An interval excluding
zero indicates agreement across these runs, not population-level
significance. Indeed, even a unanimous two-sided sign test with $n=4$ cannot
reach $p<0.05$. The intervals omit test-sample and split-to-split uncertainty,
and we make no multiplicity adjustment. Matching seed identifiers across
configurations does not make the runs identical, because the training images
and resulting optimization trajectories differ, but it avoids comparing
arbitrary seed sets and preserves a consistent pairing for differences. We
report the four seed-level observations with the interval summaries so that
the reader can distinguish a uniform shift from a mean driven by one run. The
bootstrap is used descriptively because four observations cannot support a
stable estimate of a general training-run distribution. No interval should be
interpreted as covering uncertainty from the 31 test photographs, alternative
train--validation--test partitions, new campus locations, or different object
instances. Those sources would require repeated group-aware splits or an
independent external test set, neither of which is available here.

\subsection{Experimental Design}
\label{sec:experiments}
\begin{figure}[!htb]
\centering
\includegraphics[width=0.66\textwidth]{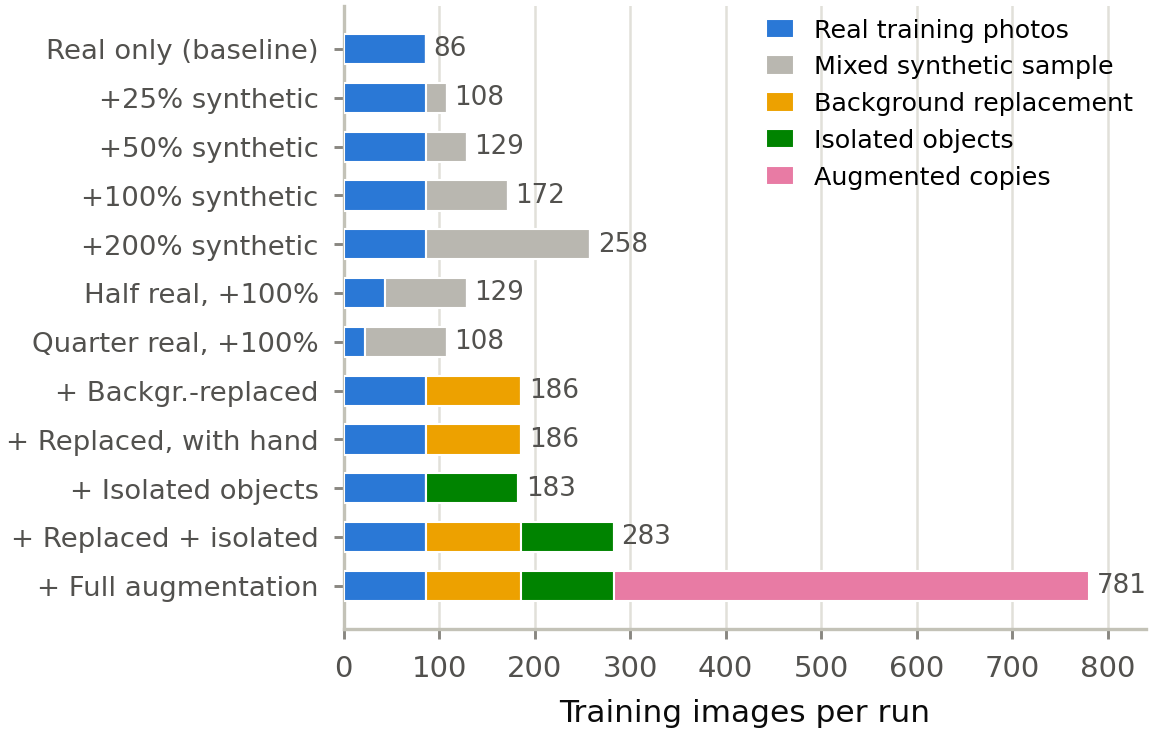}
\caption{Training configurations and image counts. The upper rows vary the
amount of added data; the lower rows change its source. ``Replaced, with
hand'' uses the background-replacement images with hand cutouts added.}
\label{fig:design}
\end{figure}

Fig.~\ref{fig:design} lays out the twelve unique joint-training
configurations. Seven rows (including the real-only baseline) vary the
\emph{amount} of added data sampled from the complete pool:
0\%, 25\%, 50\%, 100\%, and 200\% of the 86 real training photos, plus two
runs that keep synthetic data at 100\% while cutting the real photos to a half
and a quarter. Five additional rows vary the \emph{source} or the hand
intervention: they add the 100 background-replaced images, the same 100 images
with hand-and-forearm composites, the 97 isolated-object images, the two base
sources together, or all sources including the 498 transformed copies (the
full augmentation pipeline). Because the combined and
full-augmentation sets contain more images than the single-source sets,
source type and dataset size are confounded there; background-replaced versus
isolated-object (186 vs.\ 183 training images) is the size-matched comparison.
All rows retain the same detector protocol and real test set, so changes in
their reported performance can be associated with the specified training-set
construction within this experiment. They cannot be attributed to source type
alone when image count or the number of optimization exposures also changes.
For that reason, the results distinguish the controlled size-matched contrast
from the broader configurations and use the latter to describe observed
behavior rather than a pure source effect.

\paragraph{Hand-context intervention.}
Every real photograph shows an object held in a hand; none of the
background-replacement images does. We used MediaPipe Hands
~\cite{mediapipehands} to extract hand-and-forearm crops from real photographs
and alpha-composited them into the 100 background-replacement images. The
background, object pixels, class label, and bounding box remained unchanged.
The intervention therefore measures the complete composite change, including
new pixels, edges, and occlusion. The first cutout set was invalid. It
contained six crops drawn from all 148
photographs, including two from the test split. We discarded those runs and
repeated the MediaPipe scan using only the 86 training photographs. Three
usable crops remained. We then regenerated all 100 composites and retrained
the four seeds. RQ2 uses only this corrected, training-only experiment; the
discarded result is retained in Section~\ref{sec:handcontext_results} to make
the correction traceable. The three valid cutouts provide limited hand and
pose diversity, and their repeated use can itself become a recognizable
pattern. The comparison therefore asks whether this particular corrected
composite set changes detection relative to its matched no-hand source; it
does not test every way a hand could be modeled, and it does not identify
semantic context separately from the visible occlusion and boundary changes.

\paragraph{Sequential transfer learning vs.\ joint mixing.}
For three synthetic sources we compared joint mixing (training once on real
and synthetic images together, as above) against a sequential strategy:
pretrain YOLOv8n on the synthetic images alone for 100 epochs, then fine-tune
on the 86 real photos for 40 epochs at a tenfold lower learning rate. The
three sources are the complete pool (695 images), the background-replaced set
(100 images), and the isolated-object set (97 images). The pretraining stage
always hands over its final-epoch checkpoint, so no real-data feedback leaks
into pretraining. A further variant freezes the pretrained backbone and
fine-tunes only the detection head.
The transfer runs use one seed and unequal training budgets: 100 epochs for
joint mixing versus 100 synthetic-only epochs plus 40 real-only epochs for
sequential training. It can reveal source-dependent behavior, but it cannot
isolate the effect of data ordering. We consequently treat RQ3 as an
exploratory comparison rather than a matched test of curricula. A definitive
ordering experiment would equalize the number of optimization steps, repeat
each condition across seeds, and keep model-selection opportunities identical
across the joint and sequential branches.

\paragraph{Difficult-condition subsets.}
We tagged five difficult-condition subsets within the pooled 62 validation
and test photographs: the 16 darkest images, the 16 most reflective images,
and three location groups containing 17--27 images. The subsets overlap and
include validation data, so they cannot provide independent estimates of
generalization. We use them only to examine whether the primary ranking is
concentrated in particular lighting or location conditions. Pooling validation
and test photographs increases the number of examples available for this
diagnostic view, but it also means that these subset values are not additional
held-out test metrics. The darkness and reflectivity groups describe observed
image conditions, while the location groups can combine several uncontrolled
differences such as background, camera angle, and illumination. We compare the
same selected models within each group and retain the primary mAP@0.5 metric,
but we do not compute confidence intervals or claim independence across the
overlapping subsets. A pattern that agrees with the main test result can show
where failures occur; an isolated reversal can motivate further collection,
not establish location-specific superiority. These restrictions are carried
into the interpretation in Section~\ref{sec:results}.

\paragraph{Edge benchmark.}
For the edge benchmark, Ultralytics 8.4.76 exported the models to ONNX opset
12. Its built-in INT8 path~\cite{int8,krishnamoorthi} used the 31-image
validation split from the 25\%-synthetic configuration for calibration. We
timed the Ultralytics \texttt{predict()} call, which includes image loading,
preprocessing, inference, and non-maximum suppression. Each measurement used
batch size 1, confidence 0.25, IoU 0.7, 10 warm-up runs, and 100 timed runs.
The host was an AMD Ryzen Threadripper PRO 5995WX workstation (64 cores,
251~GB RAM) running Ubuntu 24.04.3 LTS. We capped execution at four threads to
approximate the core count of a Raspberry Pi 4. ONNX Runtime 1.27.0 was
installed during the analysis, although the timing log recorded only the
Ultralytics version. The cap controls available CPU parallelism, but it does
not reproduce the target board's processor architecture, clock rate, memory
bandwidth, thermal behavior, or software stack. The measurement is therefore
a four-thread workstation proxy used to compare exports under one controlled
environment, not a prediction of Raspberry Pi latency. End-to-end timing is
appropriate for the intended application because the controller must load and
preprocess a camera frame before it can act, although it should not be
compared directly with studies that report inference-only kernel time. We
record model size, latency, and whether the exported model returns detections;
a smaller file is not treated as a successful deployment result if its output
is numerically invalid.

\section{Results and Discussion}
\label{sec:results}

\subsection{The Evaluated Synthetic Mixtures Did Not Improve mAP (RQ1)}
\begin{figure}[!htb]
\centering
\includegraphics[width=0.62\textwidth]{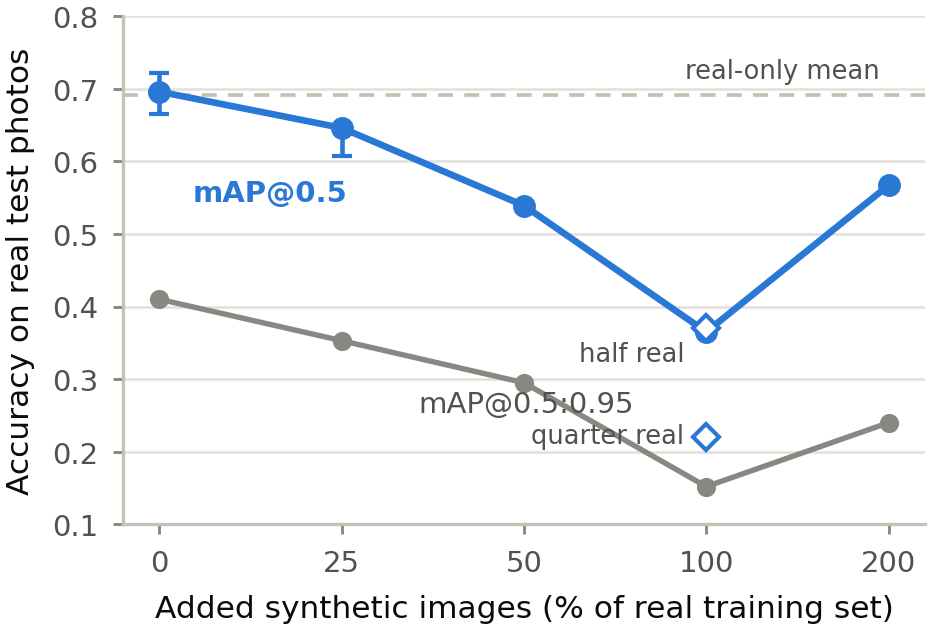}
\caption{Test-set mAP@0.5 as added images are mixed with the 86 real training
photographs. The 0\% and 25\% points show four-seed bootstrap percentile
intervals; other points use seed 2026. Open diamonds mark runs that also remove
real training photographs.}
\label{fig:mixing_curve}
\end{figure}

Real-only training produced a mean mAP@0.5 of 0.691 [0.665, 0.722], compared
with 0.633 [0.608, 0.646] after adding 25\% synthetic data. The paired
difference was $-0.059$ [$-0.114$, $-0.020$]. The remaining mixture ratios were
tested with seed 2026 only (Fig.~\ref{fig:mixing_curve}). Their response was not
monotonic: mAP@0.5 fell to 0.365 at 100\% added data, then rose to 0.567 at
200\%. At the nominal 100\% ratio, replacing half or three quarters of the real
training photographs gave 0.370 and 0.220, respectively. No mixture exceeded
the real-only reference, although only the 25\% comparison has four-seed
support. The two replacement runs further show that this added-image pool did
not compensate for removing deployment-view photographs from training. For
the 25\% condition, the interval for the matched seed-wise difference remains
below zero, so the reduction is not the product of a single unfavorable run
among the four observed seeds. The remaining ratio curve should be read more
cautiously: because each point represents seed 2026, its nonmonotonic shape
describes the tested runs rather than a stable dose--response relationship.
Even with that restriction, the curve provides no evidence for a threshold at
which additional members of this pool begin to outperform the real-only
model. The replacement results are especially relevant to data collection.
They indicate that, under this protocol, nominally increasing the number of
training files did not restore the information lost when genuine
deployment-view photographs were removed.

\subsection{Image Source Changes the Outcome (RQ1)}
\label{sec:ablation_results}
\begin{figure}[!htb]
\centering
\includegraphics[width=\textwidth]{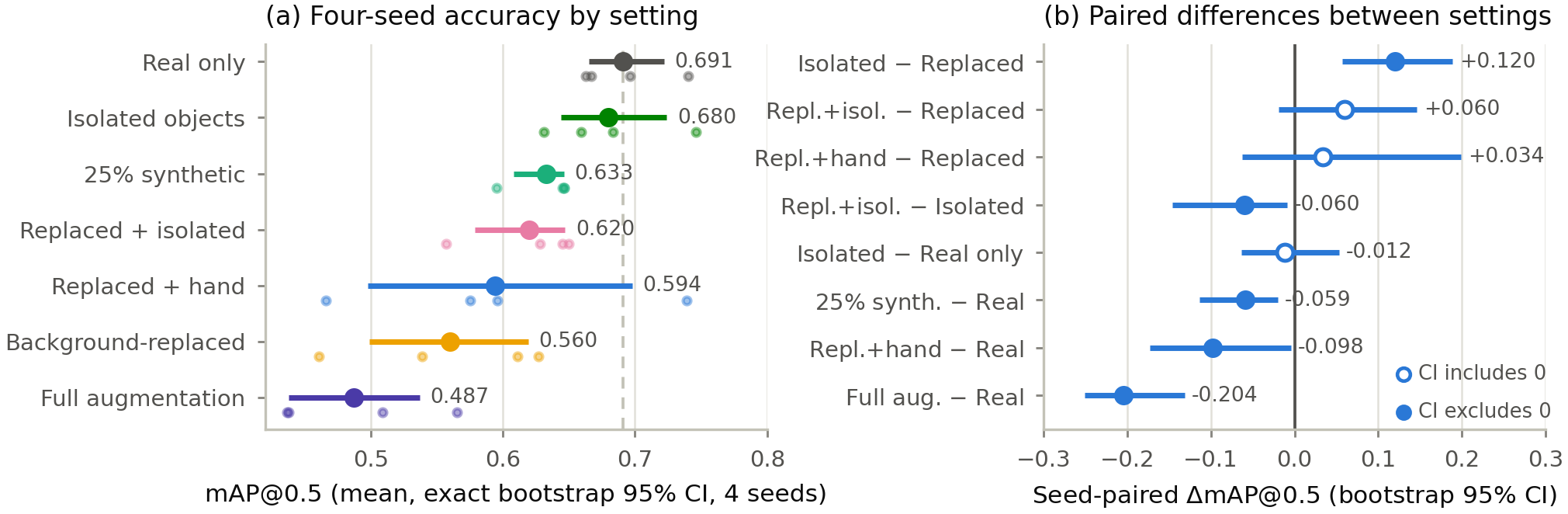}
\caption{Four-seed results. (a)~Mean mAP@0.5 and bootstrap percentile
intervals; dots show individual seeds and the dashed line gives the real-only
mean. (b)~Seed-paired differences. Filled markers denote intervals that
exclude zero.}
\label{fig:reliability}
\end{figure}

Background-replaced images produce the largest single-source reduction, from
0.691 to 0.560 [0.499,
0.619], and the highest across-seed standard deviation (0.076). For the
isolated-object images, the paired difference from real-only is $-0.012$
[$-0.064$, $+0.053$], which includes zero. Combining the two sources yields
0.620 [0.579, 0.647]; its difference from isolated objects alone is $-0.060$
[$-0.146$, $-0.009$]. The full augmentation pipeline obtains 0.487 [0.438,
0.537], with a paired difference of $-0.204$ [$-0.251$, $-0.131$] from
real-only. In the approximately size-matched comparison (186 versus 183 total
training images), isolated objects exceed background replacement by $+0.120$
[0.057, 0.189] (Fig.~\ref{fig:reliability}). Thus, the two similarly sized
training sets behave differently; image count alone does not explain the
result. The isolated-object mean lies close to the real-only mean, and its
paired interval includes zero, so this experiment does not establish either a
benefit or a loss for that source. Background replacement, by contrast, is
lower on average and more variable across seeds. Combining sources does not
recover the isolated-object result, while adding the complete transformation
pipeline produces the largest mean reduction. Because the combined and full
sets contain more images, their outcomes mix source composition with training
set size and exposure; they are not pure source ablations. The approximately
matched isolated-object and background-replacement contrast is therefore the
cleanest evidence for a source-dependent difference in RQ1. Within the limits
of four seeds and one real test split, it shows that two additions with nearly
the same file count can lead to materially different detectors.

\subsection{Corrected Hand-Composite Experiment (RQ2)}
\label{sec:handcontext_results}
\begin{figure}[!htb]
\centering
\includegraphics[width=0.62\textwidth]{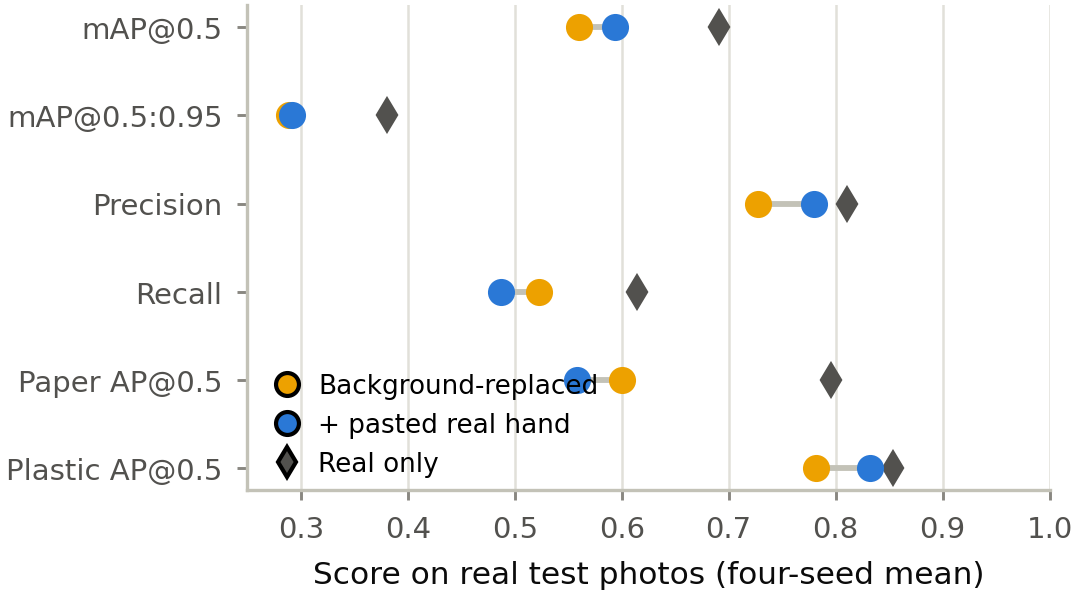}
\caption{Four-seed mean metrics for background replacement, the corrected
training-only hand composite, and real-only training. Glass is omitted because
the test set contains two examples; metal is reported in
Fig.~\ref{fig:perclass}.}
\label{fig:handcontext}
\end{figure}

\begin{figure}[!htb]
\centering
\includegraphics[width=\textwidth]{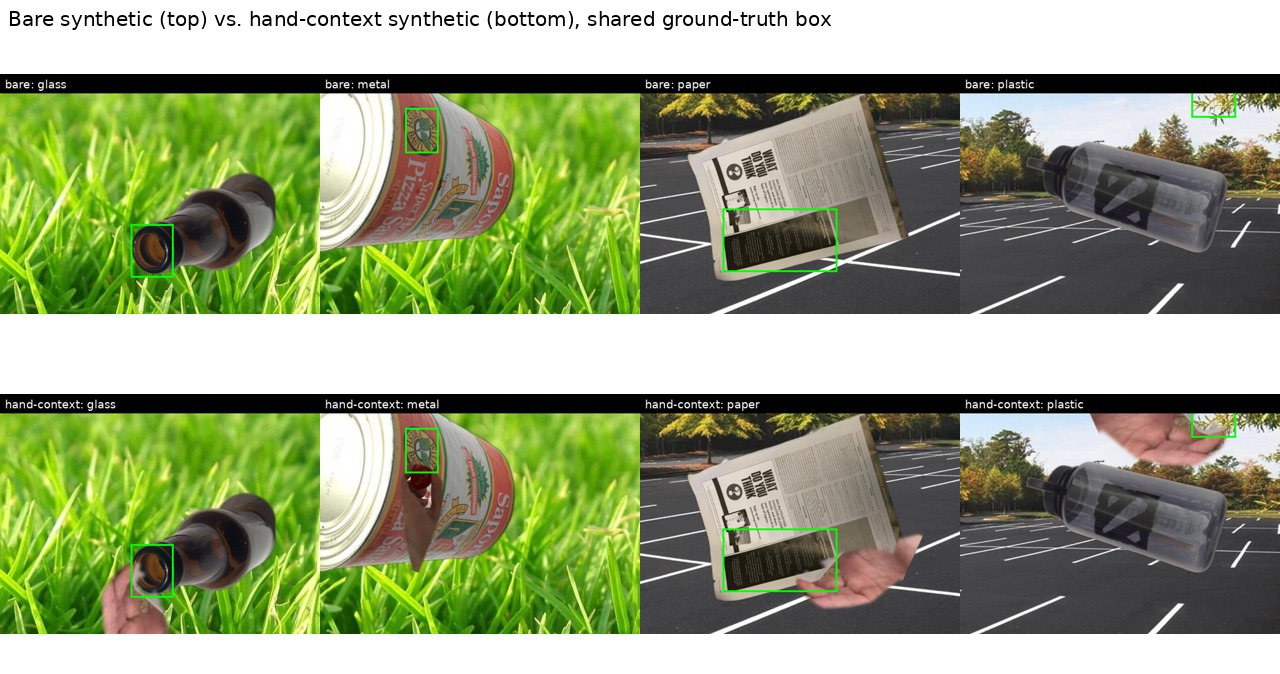}
\caption{Background-replacement images (top) and their corrected
hand-composited counterparts (bottom). All cutouts came from the 86-photo
training split. Green boxes show the unchanged ground-truth labels.}
\label{fig:handcontext_panel}
\end{figure}

The corrected hand-composite set reached a four-seed mean of 0.594 [0.498,
0.698]. Relative to the matched background-replacement set, the paired
difference was $+0.034$ [$-0.063$, $0.199$]
(Fig.~\ref{fig:reliability}b). Seed 2026 reached 0.739, whereas seeds 1--3
ranged from about 0.47 to 0.60. The interval includes zero, and the other
metrics do not move together: precision shifts slightly toward real-only
performance, while recall, paper AP@0.5, and mAP@0.5:0.95 do not
(Fig.~\ref{fig:handcontext}). For traceability, the discarded experiment used
six cutouts, two of which
came from test photographs. It produced 0.647 [0.616, 0.681] and a paired
difference of $+0.088$ [0.007, 0.183]. Those numbers are not used as evidence.
After restricting the source to training photographs, only three distinct
cutouts remained (Fig.~\ref{fig:handcontext_panel}). With three cutouts and
four seeds, the corrected experiment provides no reliable evidence that the
composite improves detection. It also changes texture, edges, and occlusion
along with hand semantics, so it cannot isolate a semantic context effect.
The contrast between the discarded and corrected estimates demonstrates why
cutout provenance matters in a small-data experiment: two test-derived crops
were enough to change the apparent strength and uncertainty of the result.
The rerun removes that leakage, but it does not solve the limited diversity of
the remaining hand library or the instability visible across seeds. RQ2 is
therefore answered narrowly: this corrected set of 100 composites did not show
a consistent advantage over its matched background-replacement source.

\subsection{Exploratory Integration-Strategy Comparison (RQ3)}
\label{sec:transfer_results}
\begin{figure}[!htb]
\centering
\includegraphics[width=0.62\textwidth]{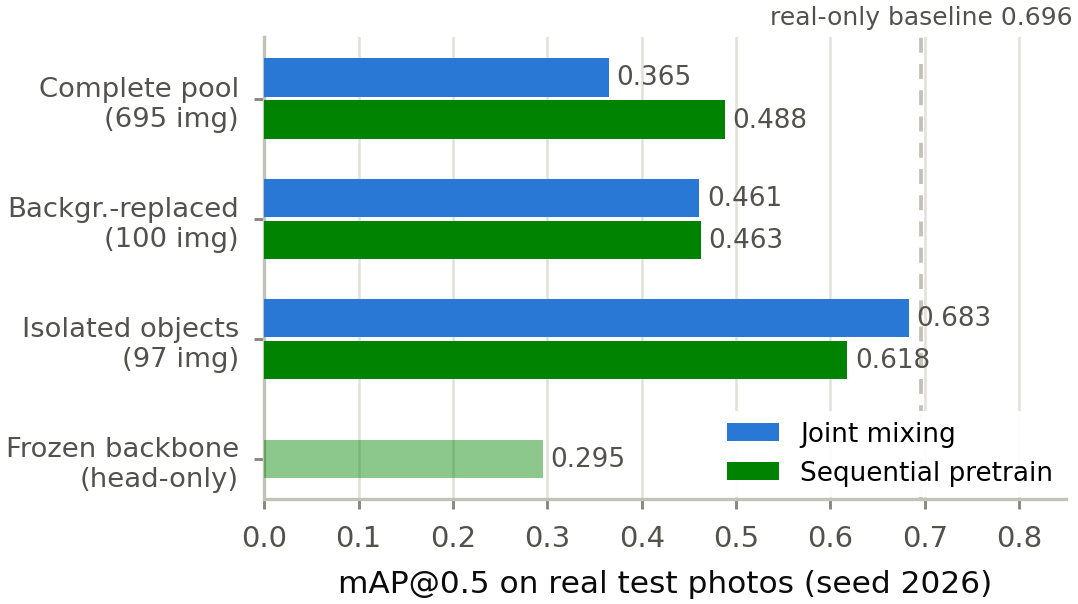}
\caption{Joint mixing and sequential pretrain--fine-tune results for seed 2026.
Sequential runs receive 140 epochs in total, compared with 100 for joint
mixing.}
\label{fig:transfer}
\end{figure}

Sequential pretraining improved the complete-pool result from 0.365 to 0.488
($+0.123$), but it reduced the isolated-object result from 0.683 to 0.618
($-0.065$). Background replacement changed little: 0.461 under joint mixing
and 0.463 under sequential training (Fig.~\ref{fig:transfer}). Fine-tuning only
the frozen model head gave 0.295. None reached the seed-2026 real-only value of
0.696. The ranking depends on the image source. We cannot attribute it to training
order, however, because each run uses one seed and sequential training receives
40 additional epochs. The frozen-backbone result is consistent with poor
transfer from this image pool~\cite{yosinski}; it does not reproduce the
real-pretrained freezing protocol of Hinterstoisser et al.~\cite{hinter}.
For the complete pool, exposure to real photographs in a separate final stage
recovers part of the joint-mixing loss, whereas the same sequence reduces the
stronger isolated-object run. The near-identical background-replacement values
show no practical separation for that source in seed 2026. These opposing
directions are more informative than a pooled statement that sequential
training either helps or hurts: they suggest that integration behavior is
source-dependent under the tested budgets. They do not demonstrate a general
curriculum effect. Equal-step, multi-seed runs would be required to separate
ordering from the extra 40 epochs and from normal training variability.

\subsection{Per-Class Results}
\label{sec:per-class}
\begin{figure}[!htb]
\centering
\includegraphics[width=0.66\textwidth]{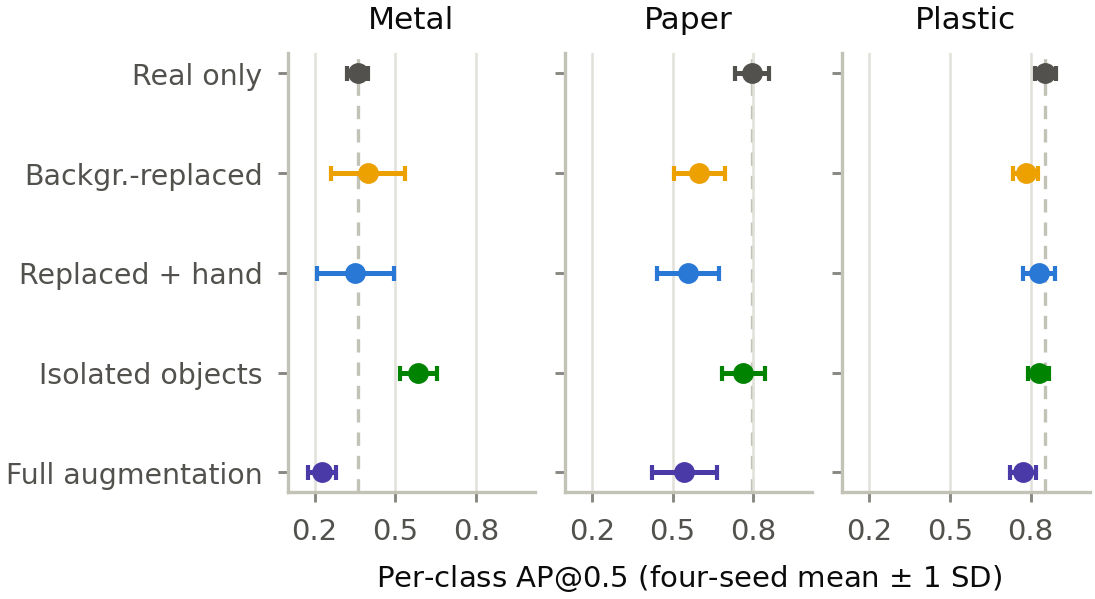}
\caption{Per-class AP@0.5 by image source (four-seed mean $\pm$ 1 SD). Dashed
lines mark the real-only means. The hand-composite values use training-only
cutouts; glass is omitted because it appears in two test photographs.}
\label{fig:perclass}
\end{figure}

Metal AP@0.5 increased from $0.360 \pm 0.039$ (mean $\pm$ one standard
deviation) with real-only training to $0.586 \pm 0.068$ with isolated-object
images, a 63\% increase in the mean (Fig.~\ref{fig:perclass}). Paper AP@0.5 was
$0.600 \pm 0.095$ with background replacement and $0.558 \pm 0.116$ with the
corrected hand composite. Plastic changed less across the displayed sources.
These class-level patterns remain tentative: scores differ by as much as 0.24
between seeds, and paired intervals were not computed. We omit glass from this
analysis because only two test photographs contain that class. The metal result
shows that an added source can help one class even when its aggregate effect is
neutral or negative, but the available experiment does not identify the visual
property responsible for that change. Likewise, the paper means do not support
a simple claim that inserting a hand restores real-scene behavior: the
corrected composite remains below background replacement for this class while
other aggregate metrics move differently. Plastic contributes the largest
number of real photographs and is correspondingly less affected by the
extreme small-sample problem that applies to glass, yet its displayed
differences are still estimated from only four runs on one split. We therefore
report the class values to expose heterogeneity that is hidden by mean mAP,
not to select a class-specific augmentation policy. More test objects and
paired, pre-specified class comparisons would be needed for that use.

\subsection{Difficult-Condition Subsets}
\begin{figure}[!htb]
\centering
\includegraphics[width=\textwidth]{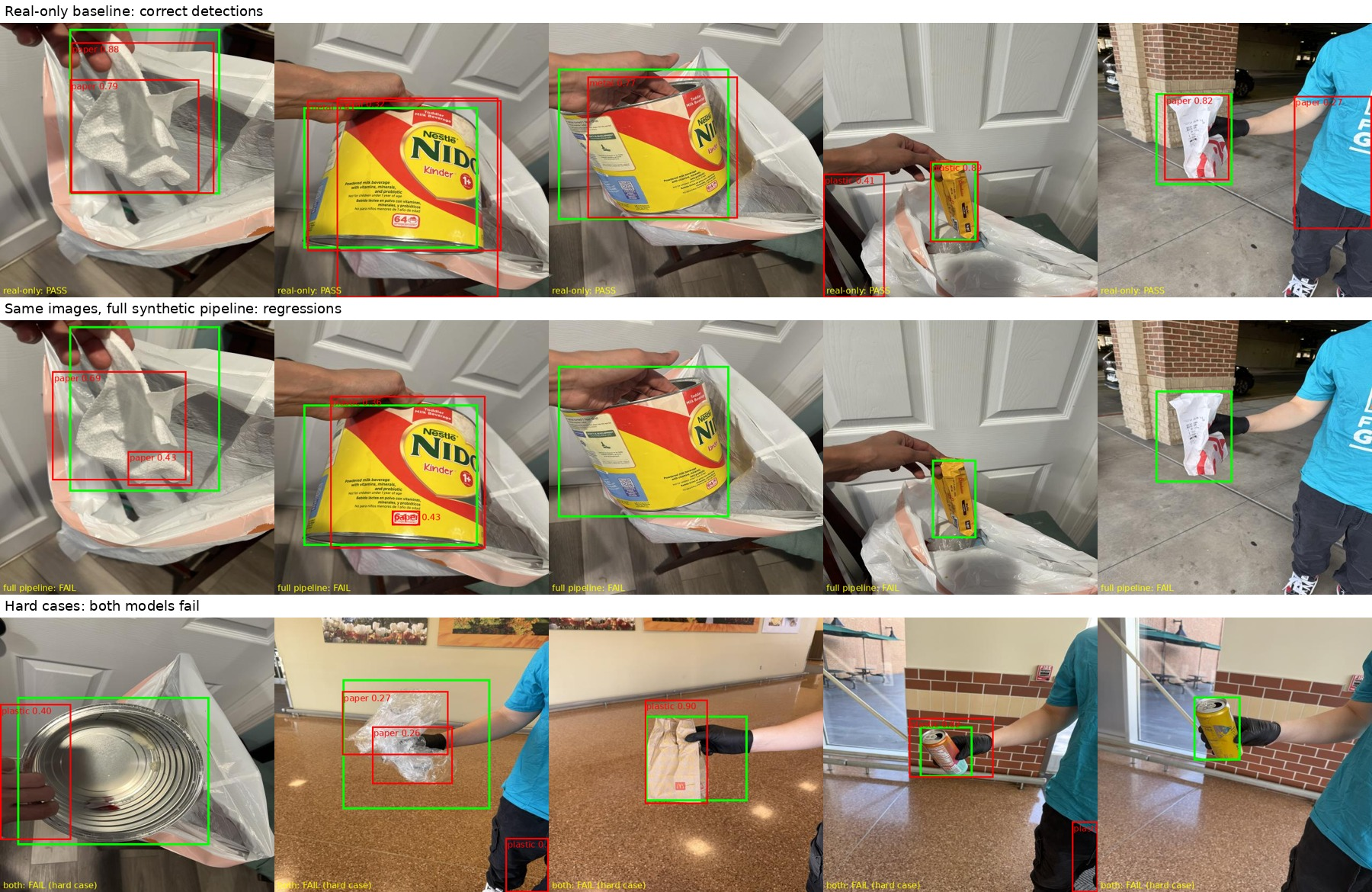}
\caption{Model predictions on real test photos (green: ground truth; red:
prediction). Row~1: photos the real-only model detects correctly. Row~2: the
same photos under the full-augmentation model, now wrong (7 regressions vs.\
1 fix out of 31 test photos). Row~3: hard photos that both models miss.}
\label{fig:qualitative_panel}
\end{figure}

\begin{figure}[!htb]
\centering
\includegraphics[width=0.62\textwidth]{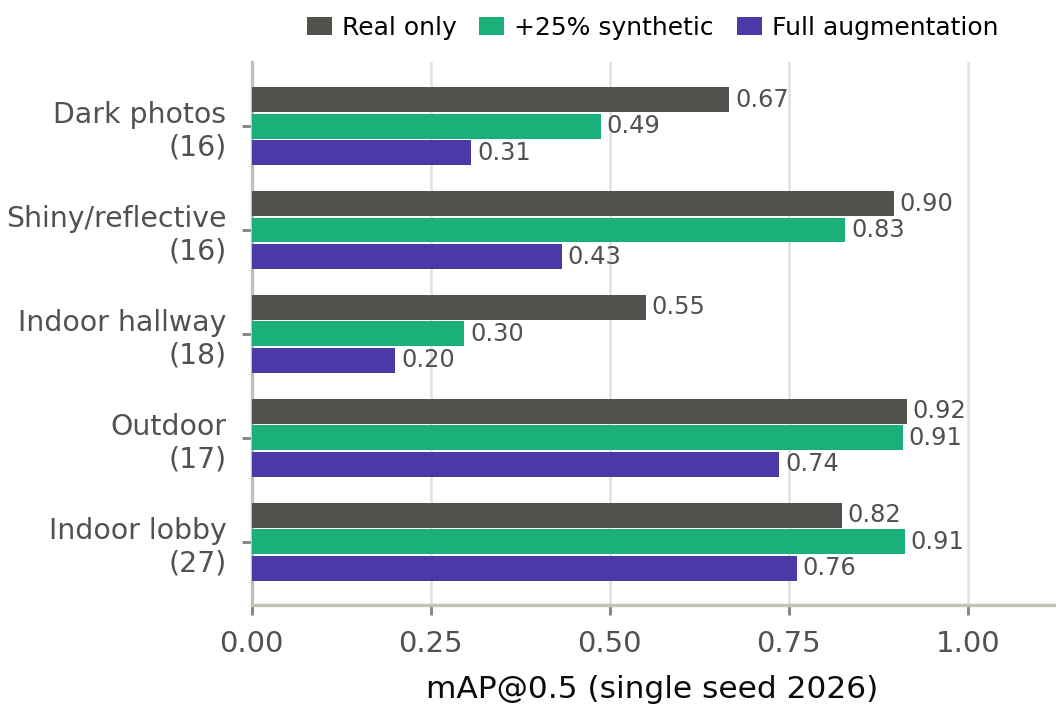}
\caption{Seed-2026 mAP@0.5 for five overlapping condition subsets pooled from
validation and test photographs. Parentheses give subset sizes.}
\label{fig:robustness}
\end{figure}

At the selected operating point, switching from the real-only model to the
full-augmentation model corrects one of the 31 test-image outcomes and changes
seven outcomes from correct to incorrect (Fig.~\ref{fig:qualitative_panel}).
This count depends on the fixed confidence and matching thresholds; mAP over
the full precision--recall curve remains the primary comparison. The count is
nevertheless relevant to the proposed bin controller, which would act on these
thresholded predictions rather than on the complete curve.
The predefined difficult-condition subsets show a similar pattern
(Fig.~\ref{fig:robustness}). Real-only training is highest or tied on four of
five subsets, while full augmentation is lowest on all five. In the exception,
the indoor-lobby subset, the 25\%-synthetic model gives 0.912 and the real-only
model 0.824. Because these overlapping subsets include validation images and
use one seed, we treat them as diagnostic examples rather than a robustness
benchmark. Taken together, the thresholded outcomes and subset scores localize
the aggregate result without changing its evidentiary basis. The seven
regressions show that the lower full-augmentation mAP is visible in individual
test cases, while the one indoor-lobby reversal shows that the ranking is not
identical in every scene group. Neither analysis supports a claim about the
frequency of such conditions on campus, and the pooled subset values cannot be
treated as independent confirmation because some photographs contribute to
more than one group. Their appropriate role is diagnostic: they identify
lighting and location conditions that should be sampled deliberately in a
larger, group-disjoint follow-up study.

\subsection{Edge Deployment}
\label{sec:results-edge}
\begin{figure}[!htb]
\centering
\includegraphics[width=0.62\textwidth]{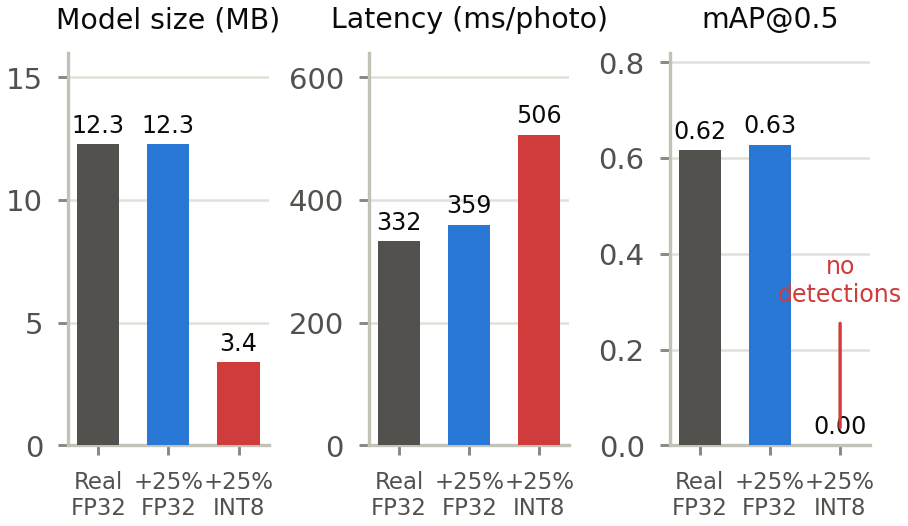}
\caption{Four-thread CPU benchmark. The FP32 models have similar size, latency,
and mAP. The smaller INT8 export is slower and produces no detections.}
\label{fig:edge}
\end{figure}

Both 32-bit floating-point (FP32) exports are 12.3~MB and require 332--359~ms per image on the
four-thread proxy, or approximately 2.8--3.0 images per second
(Fig.~\ref{fig:edge}). We did not measure the intended single-board computer.
The 3.4~MB INT8 export is slower and produces no detections. At all 8,400
detection positions, its box coordinates remain nonzero but its class scores
are zero. The failure therefore appears in the class-score path. We did not
determine whether calibration, preprocessing, quantization, an unsupported
operation, or the runtime caused it. The FP32 measurements show that the
real-only and 25\%-synthetic detectors have essentially the same deployment
footprint despite their different accuracy, as expected from their identical
architecture. Their observed throughput may be adequate for occasional
point-of-disposal feedback, but that judgment requires an end-to-end test with
the actual camera and board. The INT8 file demonstrates why size alone is an
insufficient deployment criterion: although it is approximately one quarter
of the FP32 size, it is slower in this environment and yields no usable class
prediction. Because the fault was not localized, we do not interpret the INT8
accuracy or latency as evidence against quantization in general. It is a
failed export path that must be diagnosed, recalibrated, and validated
numerically before any hardware comparison. The only supported edge result is
the controlled four-thread proxy measurement for the two functioning FP32
exports.

\section{Limitations}
\label{sec:limitations}
The 31-photo test set is small and contains only two glass examples. Every
configuration also uses the same split. Our intervals describe variation over
four training seeds, not uncertainty from the test sample or alternative data
splits, and the comparisons are not adjusted for multiplicity.
The class-stratified split could not be checked for repeated physical objects
within a source batch. Provenance and licensing records are also missing for
the isolated-object images. In addition, source, image count, and optimization
exposure are partly confounded: larger training sets produce more gradient
updates during the fixed 100 epochs. Only the background-replacement and
isolated-object sets are approximately matched in size.
The corrected hand experiment uses three training-split cutouts. That small
set provides limited visual variety, and the intervention changes pixels,
edges, and occlusion along with hand semantics. The transfer comparison uses
one seed and unequal epoch budgets. The condition subsets likewise use one
seed, overlap, and include validation photographs.
Finally, the latency measurements come from a capped desktop CPU, not the
target board. The INT8 export failed, and the responsible step in the export
or inference pipeline remains unknown. These constraints limit different parts
of the paper in different ways. The fixed split and possible object overlap
restrict generalization beyond these photographs; missing source records
restrict reproducibility and lawful redistribution of the added pool; and the
confounded training exposures restrict causal claims about why one source
performs differently. The hand, transfer, condition-subset, and edge analyses
should consequently be treated as targeted diagnostics rather than independent
confirmatory experiments. None of these limitations invalidates the recorded
comparison on the fixed test set, but together they require conservative
language about population performance, semantic context, training curricula,
and deployability.

\section{Conclusion}
\label{sec:conclusion}
No evaluated synthetic/derived configuration surpassed real-only training on
the fixed 31-photo test set. Across four seeds, background replacement reduced
mean mAP@0.5 from 0.691 to 0.560. The isolated-object mean was 0.680, and its
paired interval overlapped the real-only result. Image source mattered more
than the number of added images in the size-matched comparison.
After we removed test-split cutouts and reran the hand-composite experiment,
the corrected paired difference was $+0.034$ [$-0.063$, $0.199$]. The
corrected experiment does not establish a hand-composite benefit. The transfer
results also remain exploratory because they use one seed and unequal training
budgets. These findings answer the study questions within their stated scope.
For RQ1, the tested additions provide no evidence that increasing this pool
improves real-photo detection, and the approximately size-matched comparison
shows that source composition cannot be replaced by image count as an
explanation. For RQ2, the provenance-corrected composite is too variable to
support a reliable advantage over background replacement. For RQ3, sequential
training helps the complete pool in seed 2026 but harms the isolated-object
source, so no source-independent integration strategy can be selected. The
common implication is not that synthetic data is inherently ineffective; it
is that derived files must add deployment-relevant variation and must be
evaluated against a strong real-only baseline with enough repetition to expose
training noise.

For this deployment, collecting new object instances is a better next step
than generating more transformations of the same objects. The next experiment
should begin with group-disjoint repeated splits and a larger training-only
hand collection, then compare data sources at a fixed number of optimization
steps. It should also repeat the transfer comparison across seeds and add more
real glass examples, since the present two-image glass test subset cannot
support a class-level conclusion. Separate deployment work is needed to
benchmark the target board, diagnose the INT8 failure, and determine whether
quantization provides a useful speed--accuracy trade-off on the intended
hardware. Provenance and redistribution rights for every added image should be
documented before constructing that follow-up pool, and split assignment
should occur at the physical-object or capture-session level to prevent hidden
near duplicates. A larger hand collection should remain training-only and
should vary pose, skin appearance, sleeve, occlusion, and camera position
without borrowing pixels from validation or test photographs. Reporting both
curve-based metrics and the operating point used by the bin controller would
then connect statistical comparison with the actual disposal decision. These
steps would turn the present negative and mixed findings into a more decisive
test of which synthetic variations transfer to a campus disposal view, while
preserving the compact-model and edge-deployment constraints that motivated
the study.

\section*{Data Availability}
The authors retain the split lists, training configurations, seed-level
results, evaluation code, and plotting scripts and will provide them upon
reasonable request. No public repository was available at submission. Raw
campus photographs can be released only when institutional and privacy
requirements permit it.

\section*{Ethics and Privacy Statement}
The authors took all campus photographs and are the only people depicted; the
images show their own hands and disposal actions. The study did not recruit or
collect data from third-party participants, and no institutional review board
protocol was sought. Before submission, we checked
Figs.~\ref{fig:handcontext_panel} and~\ref{fig:qualitative_panel} for
third-party faces, license plates, and identifying building information. None
is visible beyond the general campus setting.


\begin{thebibliography}{33}

\bibitem{abdallah}
M. Abdallah, M. Abu Talib, S. Feroz, Q. Nasir, H. Abdalla, and B. Mahfood,
``Artificial intelligence applications in solid waste management: A systematic
research review,'' \emph{Waste Management}, vol. 109, pp. 231--246, 2020.

\bibitem{shorten}
C. Shorten and T. M. Khoshgoftaar, ``A survey on image data augmentation for
deep learning,'' \emph{Journal of Big Data}, vol. 6, no. 1, Art. no. 60, 2019.

\bibitem{cutpaste}
D. Dwibedi, I. Misra, and M. Hebert, ``Cut, paste and learn: Surprisingly easy
synthesis for instance detection,'' in \emph{Proc. IEEE Int. Conf. Computer
Vision (ICCV)}, 2017, pp. 1301--1310.

\bibitem{georgakis}
G. Georgakis, A. Mousavian, A. C. Berg, and J. Kosecka, ``Synthesizing
training data for object detection in indoor scenes,'' in \emph{Robotics:
Science and Systems (RSS)}, 2017.

\bibitem{ghiasi}
G. Ghiasi, Y. Cui, A. Srinivas, R. Qian, T.-Y. Lin, E. D. Cubuk, Q. V. Le, and
B. Zoph, ``Simple copy-paste is a strong data augmentation method for instance
segmentation,'' in \emph{Proc. IEEE/CVF Conf. Computer Vision and Pattern
Recognition (CVPR)}, 2021, pp. 2918--2928.

\bibitem{richter}
S. R. Richter, V. Vineet, S. Roth, and V. Koltun, ``Playing for data: Ground
truth from computer games,'' in \emph{Proc. European Conf. Computer Vision
(ECCV)}, 2016, pp. 102--118.

\bibitem{ros}
G. Ros, L. Sellart, J. Materzynska, D. Vazquez, and A. M. Lopez, ``The SYNTHIA
dataset: A large collection of synthetic images for semantic segmentation of
urban scenes,'' in \emph{Proc. IEEE Conf. Computer Vision and Pattern
Recognition (CVPR)}, 2016, pp. 3234--3243.

\bibitem{tobin}
J. Tobin, R. Fong, A. Ray, J. Schneider, W. Zaremba, and P. Abbeel, ``Domain
randomization for transferring deep neural networks from simulation to the
real world,'' in \emph{Proc. IEEE/RSJ Int. Conf. Intelligent Robots and
Systems (IROS)}, 2017, pp. 23--30.

\bibitem{domainrand}
J. Tremblay \emph{et al.}, ``Training deep networks with synthetic data:
Bridging the reality gap by domain randomization,'' in \emph{Proc. IEEE/CVF
Conf. Computer Vision and Pattern Recognition Workshops (CVPRW)}, 2018,
pp. 969--977.

\bibitem{torralba}
A. Torralba, ``Contextual priming for object detection,'' \emph{International
Journal of Computer Vision}, vol. 53, no. 2, pp. 169--191, 2003.

\bibitem{divvala}
S. K. Divvala, D. Hoiem, J. H. Hays, A. A. Efros, and M. Hebert, ``An
empirical study of context in object detection,'' in \emph{Proc. IEEE Conf.
Computer Vision and Pattern Recognition (CVPR)}, 2009, pp. 1271--1278.

\bibitem{rosenfeld}
A. Rosenfeld, R. Zemel, and J. K. Tsotsos, ``The elephant in the room,''
arXiv:1808.03305, 2018.

\bibitem{dvornik}
N. Dvornik, J. Mairal, and C. Schmid, ``Modeling visual context is key to
augmenting object detection datasets,'' in \emph{Proc. European Conf. Computer
Vision (ECCV)}, 2018, pp. 364--380.

\bibitem{trashnet}
G. Thung and M. Yang, ``Classification of trash for recyclability status,''
CS229 Course Report, Stanford University, 2016.

\bibitem{taco}
P. F. Proen\c{c}a and P. Sim\~{o}es, ``TACO: Trash annotations in context for
litter detection,'' arXiv:2003.06975, 2020.

\bibitem{warp}
D. Yudin \emph{et al.}, ``Hierarchical waste detection with weakly supervised
segmentation in images from recycling plants,'' \emph{Engineering Applications
of Artificial Intelligence}, vol. 128, Art. no. 107542, 2024.

\bibitem{zerowaste}
D. Bashkirova \emph{et al.}, ``ZeroWaste dataset: Towards deformable object
segmentation in cluttered scenes,'' in \emph{Proc. IEEE/CVF Conf. Computer
Vision and Pattern Recognition (CVPR)}, 2022, pp. 21147--21157.

\bibitem{jin2023}
S. Jin, Z. Yang, G. Kr\'{o}lczyk, X. Liu, P. Gardoni, and Z. Li, ``Garbage
detection and classification using a new deep learning-based machine vision
system as a tool for sustainable waste recycling,'' \emph{Waste Management},
vol. 162, pp. 123--130, 2023.

\bibitem{fasterrcnn}
S. Ren, K. He, R. Girshick, and J. Sun, ``Faster R-CNN: Towards real-time
object detection with region proposal networks,'' in \emph{Advances in Neural
Information Processing Systems (NeurIPS)}, 2015, pp. 91--99.

\bibitem{yolo}
J. Redmon, S. Divvala, R. Girshick, and A. Farhadi, ``You only look once:
Unified, real-time object detection,'' in \emph{Proc. IEEE Conf. Computer
Vision and Pattern Recognition (CVPR)}, 2016, pp. 779--788.

\bibitem{ssd}
W. Liu, D. Anguelov, D. Erhan, C. Szegedy, S. Reed, C.-Y. Fu, and A. C. Berg,
``SSD: Single shot multibox detector,'' in \emph{Proc. European Conf. Computer
Vision (ECCV)}, 2016, pp. 21--37.

\bibitem{mobilenetv2}
M. Sandler, A. Howard, M. Zhu, A. Zhmoginov, and L.-C. Chen, ``MobileNetV2:
Inverted residuals and linear bottlenecks,'' in \emph{Proc. IEEE/CVF Conf.
Computer Vision and Pattern Recognition (CVPR)}, 2018, pp. 4510--4520.

\bibitem{yolov8}
G. Jocher, A. Chaurasia, and J. Qiu, ``Ultralytics YOLOv8,'' version 8.0,
2023. [Online]. Available: \url{https://github.com/ultralytics/ultralytics}

\bibitem{int8}
B. Jacob \emph{et al.}, ``Quantization and training of neural networks for
efficient integer-arithmetic-only inference,'' in \emph{Proc. IEEE/CVF Conf.
Computer Vision and Pattern Recognition (CVPR)}, 2018, pp. 2704--2713.

\bibitem{krishnamoorthi}
R. Krishnamoorthi, ``Quantizing deep convolutional networks for efficient
inference: A whitepaper,'' arXiv:1806.08342, 2018.

\bibitem{hinter}
S. Hinterstoisser, V. Lepetit, P. Wohlhart, and K. Konolige, ``On pre-trained
image features and synthetic images for deep learning,'' in \emph{Proc.
European Conf. Computer Vision Workshops (ECCVW)}, 2018, pp. 682--697.

\bibitem{dietterich}
T. G. Dietterich, ``Approximate statistical tests for comparing supervised
classification learning algorithms,'' \emph{Neural Computation}, vol. 10,
no. 7, pp. 1895--1923, 1998.

\bibitem{bouthillier}
X. Bouthillier \emph{et al.}, ``Accounting for variance in machine learning
benchmarks,'' in \emph{Proceedings of Machine Learning and Systems (MLSys)},
vol. 3, 2021, pp. 747--769.

\bibitem{bootstrap}
B. Efron and R. J. Tibshirani, \emph{An Introduction to the Bootstrap}.
New York, NY, USA: Chapman \& Hall, 1993.

\bibitem{coco}
T.-Y. Lin \emph{et al.}, ``Microsoft COCO: Common objects in context,'' in
\emph{Proc. European Conf. Computer Vision (ECCV)}, 2014, pp. 740--755.

\bibitem{voc}
M. Everingham, L. Van Gool, C. K. I. Williams, J. Winn, and A. Zisserman,
``The PASCAL visual object classes (VOC) challenge,'' \emph{International
Journal of Computer Vision}, vol. 88, no. 2, pp. 303--338, 2010.

\bibitem{mediapipehands}
F. Zhang, V. Bazarevsky, A. Vakunov, A. Tkachenka, G. Sung, C.-L. Chang, and
M. Grundmann, ``MediaPipe Hands: On-device real-time hand tracking,''
arXiv:2006.10214, 2020.

\bibitem{yosinski}
J. Yosinski, J. Clune, Y. Bengio, and H. Lipson, ``How transferable are
features in deep neural networks?'' in \emph{Advances in Neural Information
Processing Systems (NeurIPS)}, 2014, pp. 3320--3328.

\end{thebibliography}
\end{document}